\documentclass{amsart}

\usepackage{graphics}
\usepackage{a4wide}
\usepackage{graphicx, tikz}

\usepackage{hyperref}
\usepackage{color}
\usepackage{url}
\usepackage{enumerate}
\usepackage{amsmath,amsthm,amssymb}
\usepackage{caption}
\usepackage{subcaption}

\usepackage{lscape}

\usepackage{lscape}

\newtheorem{theorem}{Theorem}[section]

\newtheorem*{lem}{Lemma}

\newtheorem{observe}[theorem]{Observation}

\newtheorem{remark1}[theorem]{Remark}

\newcommand{\diam}{\operatorname{diam}}

\newcommand{\s}[1]{\textcolor{black}{#1}}
\newcommand{\st}[1]{\textcolor{black}{#1}}

\begin{document}

\title[]{Stochastic Neighbor Embedding \\separates well-separated clusters}

\author[]{Uri Shaham}
\address{Department of Statistics, Yale University, New Haven, CT 06511, USA}
\email{uri.shaham@yale.edu }

\author[]{Stefan Steinerberger}
\address{Department of Mathematics, Yale University, New Haven, CT 06511, USA}
\email{stefan.steinerberger@yale.edu}

\begin{abstract}
Stochastic Neighbor Embedding and its variants are widely used dimensionality reduction techniques -- despite
their popularity, no theoretical results are known. We prove that the optimal SNE embedding of well-separated clusters
from high dimensions to \s{any Euclidean space}~$\mathbb{R}^d$ manages to successfully separate the clusters in a quantitative way. 
\s{The result also applies to a larger family of methods including a variant of t-SNE.}
\end{abstract}

\date{}


\maketitle

\vspace{-20pt}
 
\section{Introduction and main result} \label{sec:intro}
\subsection{Introduction.}
Stochastic Neighbor Embedding~\cite{hinton2002stochastic} and its variations are heavily used in dimensionality reduction (mainly for the purpose of visualization). The most widely used algorithm in this family is undoubtedly
 $t$-distributed Stochastic Neighborhood Embedding~\cite{maaten2008visualizing}, however, there are several others: we mention symmetrized versions~\cite{CookSMH07}, a variant with emphasis on heavy tails \cite{yang2009heavy} and an algorithm based on triplet comparison \cite{MaatenW12}. We also regard \cite{lawrence} and \cite{semantic} as related in spirit and refer to \cite{hubs} for a more detailed investigation of
clusters.
Despite the popularity of these methods, we are not aware of any theoretical result that would explain the mechanism that underlies the success of SNE \s{and its variants}.
The purpose of our paper is to make a modest step in that direction and to show that at least some theoretical analysis is possible: we prove that \s{an entire family of these methods, including SNE and a variant of t-SNE,} successfully maps well-separated disjoint clusters from high dimensions to the real line so as to approximately preserve the clustering. 
 We hope that this will inspire further theoretical work on stochastic neighborhood embedding. The problem is also interesting from a purely mathematical point of view: it is similar in spirit to classical
problems regarding the optimal arrangement of points on a sphere (or manifold) to minimize the energy arising from pairwise interaction under some potential (e.g. \cite{saff}) and perhaps
some of the techniques carry over.

\subsection{SNE and our Setup} \label{sec:preliminaries} We introduce the stochastic neighborhood embedding (SNE) functional and fix notation.
Given set of $k$ points, \s{ $\mathcal{X} = \{x_1,\ldots,x_k\} \subset \mathbb{R}^D$}, stochastic neighborhood embedding searches for \s{$\psi = \{\psi_1,\ldots,\psi_k\} \subset \mathbb{R}^d $ (usually $d \in \left\{2, 3\right\}$)} that minimizes the loss
$$
L(\psi_1,\ldots,\psi_k) = -\sum_{i=1}^{k}\sum_{j = 1 \atop j\neq i}^{k}p(x_i,x_j)\log q(\psi_i,\psi_j), 
$$
where the functions $p$ and $q$ are given by
$$
p(x_i,x_j) = \frac{\exp(-\|x_i - x_j\|^2/2\sigma_i^2)}{\sum_{\ell \neq i}{\exp(-\|x_i - x_{\ell}\|^2/2\sigma_i^2)}}  \qquad \mbox{and} \qquad q(\psi_i,\psi_j)  = \frac{\exp(-\|\psi_i-\psi_j\|^2)}{\sum_{\ell \neq i}{\exp(-\|\psi_i - \psi_{\ell}\|^2)}},
$$

The proper scales $\sigma_i$ are set either by hand, or according to various guidelines (which will not be of importance in our approach). Our idealized setup of well-separated clusters is as follows:
\begin{itemize}
\item We are given $n$ sets \s{$B_1, \dots, B_n \subset \mathbb{R}^D$} each of which contains exactly $a \geq 2$ points.
\item We assume furthermore that each of the sets is localized, i.e., 
\begin{equation}
\diam(B_i) \leq 1 \notag
\end{equation}
and that the sets are separated so that for all $i \neq j$
$$
\min_{x \in B_i, y \in B_j}{\|x -y\|} \geq \sqrt{ 5\log{n}}.
$$
\end{itemize}

\subsection{Quality of an Embedding.} Ideally, we would like that clusters $B_1, \dots, B_n$ are mapped to clusters \s{in the embedding space}
and will now, for any embedding $\psi$, define a measure $Q(\psi)$ to quantify how well that goal is achieved.
We define $Q(\psi)$ as the expectation of a random variable $N$, generated via the following process:
\begin{enumerate}
\item Pick one of the clusters uniformly at random.
\item \s{Pick two different elements $x_{}, y$ uniformly at random from that cluster.}
\item \st{  Let $B(\psi(x),\| \psi(y) - \psi(x)\|) \subset \mathbb{R}^d$ denote the ball centered at $\psi(x)$ touching $\psi(y)$.}
\item \s{  Compute the logarithm of the number of points that are being mapped to the ball
\begin{equation}
N = \log{\left(| \psi \cap B(\psi(x),\| \psi(y) - \psi(x)\|)  |\right)}.\notag
\end{equation}
}
\end{enumerate}
We define the quality of an embedding $Q(\psi)$ as the expected value of $N$. Formally,
$$
Q(\psi) := \mathbb{E}(N) = \frac{1}{n} \sum_{m=1}^{n}  \frac{1}{a(a-1)}\sum_{(x_{},y_{})\in B_m}\log\left(|\psi \cap B(\psi(x),\| \psi(y) - \psi(x)\|)  |  \right),\label{eq:Q}
$$
\s{where $ B(\psi(x),\| \psi(y) - \psi(x)\|) $ is the ball  $\psi(x)$ with radius $r = \| \psi(y) - \psi(x) \|$. }Note
that this notion is more topological than metric: it does not measure actual distances between points but is merely concerned with the proper ordering.

\begin{center}
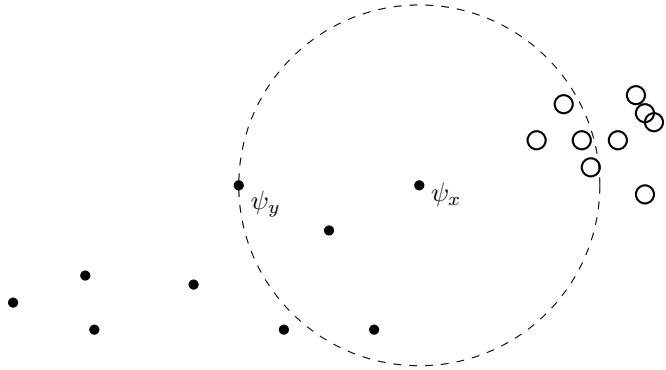
\begin{figure}[h!]
\begin{tikzpicture}[scale=1.2]
\filldraw (0.3,0.3) circle (0.05cm);
\filldraw (0.4,-0.3) circle (0.05cm);
\filldraw (-0.5,0) circle (0.05cm);
\filldraw (1.5,0.2) circle (0.05cm);
\filldraw (2.5,-0.3) circle (0.05cm);
\filldraw (2,1.3) circle (0.05cm);
\filldraw (3.5,-0.3) circle (0.05cm);
\filldraw (3,0.8) circle (0.05cm);
\filldraw (4,1.3) circle (0.05cm);
\draw [dashed] (4,1.3) circle (2.0cm);
\node at (4.3, 1.2) {$\psi_x$};
\node at (2.3, 1.1) {$\psi_y$};
\draw [ thick] (5.3,0.3+1.5) circle (0.1cm);
\draw  [ thick] (5.6,0.7+1.5) circle (0.1cm);
\draw  [ thick] (6.5,-0.3+1.5) circle (0.1cm);
\draw [ thick] (5.8,0.3+1.5) circle (0.1cm);
\draw [ thick] (5.9,0+1.5) circle (0.1cm);
\draw [ thick] (6.2,0.3+1.5) circle (0.1cm);
\draw [ thick] (6.4,0.8+1.5) circle (0.1cm);
\draw [ thick] (6.5, 0.6+1.5) circle (0.1cm);
\draw [ thick] (6.6, 0.6+1.4) circle (0.1cm);
\end{tikzpicture}
\caption{\s{An embedding of clusters into $\mathbb{R}^2$: two randomly chosen points from the left cluster may well give rise to balls containing points from the other cluster (which increases $Q$) while
two randomly chosen points from the right cluster can only give rise to balls containing elements from the right cluster.}.}
\label{fig:embedding}
\end{figure}
\end{center}

The motivation is obvious: 
in order for an embedding to be good, we would like points from the same cluster to be mapped to nearby regions
of space in an unambiguous way, such that there are not many points between them (except possibly other points from the same cluster).
\s{We start by showing that the quality $Q(\psi)$ cannot be arbitrarily small because of the randomness in the selection process. The following
Lemma provides a lower bound on $Q$ and is independent of the number of clusters $n$, the input dimension $D$ and the output dimension $d$.}
\s{
\begin{lem}\label{lemma:notSmall}
Under the assumptions above, any embedding $\psi$ satisfies
$$ Q(\psi) \geq \log{(a)} -1.$$
Moreover, there exists an embedding $\psi$ for which $Q(\psi) \leq \log{a}.$
\end{lem}
}

\subsection{Main result} Our main result states that \s{if the input data is already well separated (in the sense specified above), then the optimal SNE embedding $\psi^*$ into \textit{any} dimension $d$ preserves
the cluster structure in the sense that its quality $Q(\psi^*)$ is as small as that of perfectly clustered embedding up to universal constants.}

\begin{theorem}[Embedding guarantee for SNE]\label{thm:R}  
Let $\psi^*$ be an embedding attaining a global minimum of the SNE loss functional with scales $\sigma_i = 2^{-1/2}$ under the assumptions above. Then 
\st{
\begin{equation}
Q(\psi^*) \leq  200\log{(2a)}.\notag
\end{equation}
}
\end{theorem}
\s{One} remarkable aspect is that this upper bound is completely independent of the number of clusters \s{$n$} (of course, a large number of clusters requires a larger degree
of separation between any two of them -- some condition like this seems necessary for SNE to successfully maintain cluster structure).
The upper bound could be improved to  \st{$Q(\psi^*) \leq 15\log{(2a)}$} if we knew that $n \gg a \gg 1$.
This result, while not giving any sort of pointwise guarantees, assures us that 'most' clusters are essentially mapped to highly localized regions in space and that the number of mismatches is small.
We also observe that the conditions in the formal setup as well as the constants in the result do not depend on the input dimension \s{$D$} nor on the output dimension \s{$d$}. \\

\s{We also mention that, while the result is independent of the output dimension $d$, it seems to be strongest for $d=1$ because, as the output dimension increases, there is 'more room'
and this seems like it would simplify the embedding problem. We also emphasize that there is one way in which a larger output dimension $d \geq 2$ helps: it allows us to relax the separation condition
 $$
\min_{x \in B_i, y \in B_j}{\|x -y\|} \geq \sqrt{ 5\log{n}} \qquad \mbox{to} \qquad \min_{x \in B_i, y \in B_j}{\|x -y\|} \geq \sqrt{ \left( 1 + \frac{2}{d} \right) \log{n}}.
$$
}

\begin{center}
\begin{figure}[h!]
\begin{tikzpicture}[scale=1]
\node (whitehead) at (0,0)
    {\includegraphics[width=\textwidth]{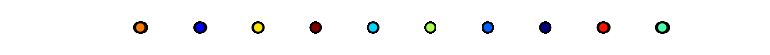}};
\draw [thick] (-5, 0) -- (6,0);
\end{tikzpicture}
\caption{The typical result of embedding 10 well-separated clusters: they are being mapped to 10 well-separated regions on the real line.}
\label{fig:figo}
\end{figure}
\end{center}

\textit{Example.}
This example consists 10 Gaussian clusters in 100 dimensions that are chosen to be very well separated --  we picked 100 points randomly from each cluster. 
Figure~\ref{fig:figo} shows the embedding of a typical realization of the data into the real line obtained using SNE: clusters remain separable, no mismatches occur. 
For the computation we used our own implementation of SNE in TensorFlow \cite{abadi2016tensorflow} using Adam~\cite{kingma2014adam} to perform the optimization.
A sketch of a typical outcome is given in Fig \ref{fig:figo}, \s{the embedding is close to perfect.}
\vspace{-15pt}
\vspace{-15pt}

\subsection{Extensions \s{and t-SNE}} \s{Our argument is not
very sensitive with respect to the kernel $q(\cdot, \cdot)$ being used: more precisely, if we replace $q(\cdot, \cdot)$ by
$$  q_2(\psi_i,\psi_j)  = \frac{f(\|\psi_i-\psi_j\|)}{\sum_{\ell \neq i}{f(\|\psi_i - \psi_{\ell}\|)}},
$$
then a result of the form $Q(\psi^*) \leq C \log{a}$, for some constant $C=C_f$ only depending on $f$, still holds whenever
 $f:\mathbb{R}_{\geq 0} \rightarrow \mathbb{R}_{>0}$ is monotonically decreasing, $f(x) \geq \alpha e^{- \beta x^2}$ for some $\alpha, \beta > 0$ and
$$ \sum_{k=1}^{\infty}{f(k)} < \infty.$$
Example of functions that could also be used are $f(x) = (1+x^2)^{-1}$ or $f(x) = \exp{(-\alpha x)}$ for $\alpha>0$. In particular, the first example gives rise to
a variant of t-SNE that is known to work very well in practice \cite{laurens}. Classical t-SNE uses a different overall normalization that makes the problem slightly
more nonlocal and keeps our argument from being applicable.}

\subsection{Open questions.} We believe that gaining a deeper understanding of SNE is not only useful because of its wide applicability and use but should also be of intrinsic
mathematical interest:  a very natural question is what happens in our model setup when the separation condition on the clusters stops being applicable (i.e. the kernel $p(\cdot, \cdot)$
has scale $\sigma = 1$, the clusters satisfy $\diam(B_i) \leq 1$ but the distance between any two clusters is smaller than $\sqrt{5 \log{n}}$). 
\vspace{-10pt}
\begin{figure}[h!]
\begin{tikzpicture}
\node (whitehead) at (-4,0)
    {\includegraphics[width=0.4\textwidth]{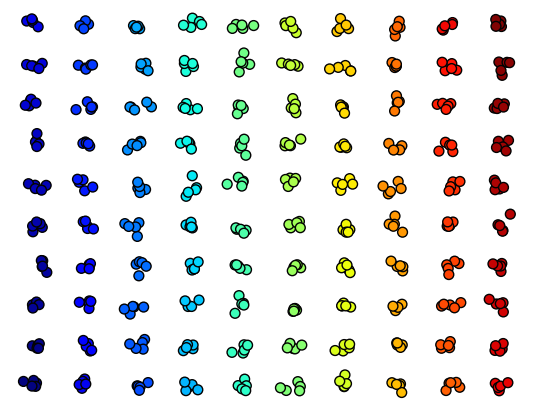}};
\node (whitehead) at (3.5,2)
    {\includegraphics[width=0.5\textwidth]{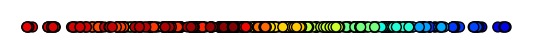}};
\node (whitehead) at (3.5,0)
    {\includegraphics[width=0.5\textwidth]{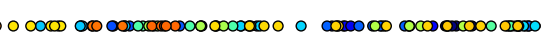}};
\node (whitehead) at (3.5,-2)
    {\includegraphics[width=0.5\textwidth]{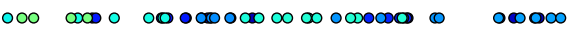}};
\end{tikzpicture}
\caption{A point set in $\mathbb{R}^2$ violating the separation condition (left), the SNE embedding in $\mathbb{R}$ (right, top) and zoomed in pictures (right, middle and bottom).}
 \label{fig:microstructure}
\end{figure}

Numerically, we observe (see Fig. \ref{fig:microstructure}) that as the separation condition is violated, the clustered embedding breaks down
and clusters start getting mixed. It is interesting to observe that an overall structure is still preserved and even on a local scale, many clusters still preserve some sort of localized structure: we see
imperfect clustering with mismatches but an overall preservation of structures on a macroscopic scale. 
A better understanding might hint at some of the underlying mechanisms of SNE and could be of great interest.
There are many other open problems: is it true that SNE performs flawlessly under our assumptions above or, more precisely, is $Q(\psi^*)$ always attaining the smallest possible value?

\vspace{0pt}

\section{Proofs} \label{sec:proofs}

\subsection{Proof of the Lemma}
\begin{proof}  \s{It is clear that assuming $\psi$ to map all elements from the same clusters to nearby regions while successfully separating the embedded clusters can only decrease the quality.
Pick now a random element $\psi(x)$ and study the distances of $\psi(y)$ (with $y$ from the same cluster) to $\psi(x)$. It is again clear that assuming all these distances to be unique will further
decrease the quality (because repeated distances occur in more balls). The simple ordering on the real line then implies
$$ Q(\psi) \geq \frac{1}{a-1} \sum_{i=1}^{a-1}{\log{(i+1)}}$$
and equality is achieved for embeddings preserving the cluster structure.
The classical comparison trick for monotone functions shows that for all $a \geq 2$
\begin{align*} \frac{1}{a-1} \sum_{i=1}^{a-1}{\log{(i+1)}} &\geq \frac{1}{a-1} \int_{1}^{a}{\log{x} dx} \\
&= \frac{1}{a-1} \left(a\log{a} - a + 1\right) \geq  \log{a} - 1.
\end{align*}
It is easy to see that for a 'perfect' embedding $\psi$, we have
$$Q(\psi) = \frac{1}{a-1} \sum_{i=1}^{a-1}{\log{(i+1)}} \leq \log{a}.$$
}
\end{proof}

\subsection{Proof of the Main Theorem}
\begin{proof} Note that all the scales are set to $\sigma_i = 1/\sqrt{2}$. We will now bound the term
$$p(x_i,x_j) = \frac{\exp(-\|x_i - x_j\|^2)}{\sum_{\ell \neq i}{\exp(-\|x_i - x_{\ell}\|^2)}}$$
from above. If $x_i$ and $x_j$ are in the same cluster, then we can find at least $a-1$ other points in the cluster that are at distance at most 1 from $x_i$ and therefore and
$$ \frac{\exp(-\|x_i - x_j\|^2)}{\sum_{\ell \neq i}{\exp(-\|x_i - x_{\ell}\|^2)}} \leq \frac{1}{\sum_{\ell \neq i}{\exp(-\|x_i - x_{\ell}\|^2)}} \leq \frac{1}{(a-1) e^{-1}} \leq \frac{2e}{a}.$$
The separation condition on the clusters
$$ c:=  \min_{x \in B_i, y \in B_j \atop i \neq j}{\|x -y\|} \geq \sqrt{5 \log{n}}$$
 implies altogether, by repeating the argument, that
$$
p(x_i,x_j) \leq \begin{cases} 2e/a \qquad &\mbox{if}~x_i~ \mbox{and}~ x_j~\mbox{are both in the same cluster} \\ 2e^{1-c^2}/a \qquad &\mbox{otherwise.} \end{cases}
$$
We start by computing an upper bound on the loss of the optimal SNE embedding: we do this by computing the loss for one particular example and using that as an upper bound. Our example is given by mapping
the cluster $B_i$ to \s{$(i,0,0,\dots,0) \in \mathbb{R}^d$} for every $1 \leq i \leq n$, we now compute its loss. It is easy to see that
$$ 
q(\psi_i,\psi_j) = \frac{\exp\left(-\|\psi_i - \psi_j\|^2 \right)}{ \sum_{k \neq i} \exp\left(-\|\psi_i - \psi_k\|^2 \right)} \geq \frac{\exp\left(-\|\psi_i - \psi_j\|^2 \right)}{ \sum_{k=-\infty}^{\infty}{a e^{-k^2}}} \geq  \frac{\exp\left(-\|\psi_i - \psi_j\|^2 \right)}{ 2a}
$$ 
and therefore, as a consequence,
$$
q(\psi_i,\psi_j) \geq \begin{cases} 1/(2a) \qquad &\mbox{if}~\psi_i~ \mbox{and}~ \psi_j~\mbox{are both in the same cluster} \\  e^{-n^2}/(2a) \qquad &\mbox{otherwise.} \end{cases}
$$

We obtain that the smallest loss $L(\psi^*)$ is bounded from above by the loss of this particular embedding
$$
L(\psi^*) \leq \sum_i\sum_{j\neq i}{ p(x_i,x_j) \log{\frac{1}{q(\psi_i,\psi_j)}}} \qquad \mbox{which we now compute.}$$
For any fixed point (out of the total of $an$ points), there are $a-1$ points in the same cluster and $(n-1)a$ points in other clusters.
Therefore, for any fixed $x_i$
\begin{align*}
 \sum_{j\neq i}{ p(x_i,x_j) \log{\frac{1}{q(\psi_i,\psi_j)}}} &\leq \frac{2e}{a}(\log{2a}) a + \left(2\frac{e^{1-c^2}}{a} (n^2 + \log{(2a)})\right)na \\
&= 2e\log{2a}+2e^{1-c^2} (n^2 + \log{(2a)})n.
\end{align*}
It is easy to see that under the assumption on $c$
$$ 2e^{1-c^2} (n^2 + \log{(2a)})n \leq \frac{2e}{n^2} + 2e\frac{\log{(2a)}}{n^4}.$$
and then, by summing over all points,
$$ L(\psi^*) \leq n a 2e \log{2a} + \frac{2ea}{n} + \frac{2ea\log{(2a)}}{n^3} \leq 6e na \log{(2a)}.$$
At the same time we see that, for any $x_i, x_j$ in the same cluster,
$$ p(x_i, x_j) =  \frac{\exp(-\|x_i - x_j\|^2)}{  \sum_{k \neq i}{\exp(-\|x_i - x_k\|^2)}} \geq \frac{e^{-1}}{a + na e^{-c^2}} \geq \frac{e^{-1}}{a+ a n^{-4}} \geq \frac{1}{6a}.$$
On the other hand, for any given embedding $\psi$,
\begin{align*}
L(\psi) = \sum_i\sum_{j\neq i}{ p(x_i,x_j) \log{\frac{1}{q(\psi_i,\psi_j)}}} \geq \sum_{m=1}^{n}{ \sum_{(x_i,x_j)\in B_m}  p(x_i,x_j) \log{\frac{1}{q(\psi_i,\psi_j)}}}.
\end{align*}

Using that for $x_i, x_j$ in the same cluster $p(x_i, x_j) \geq 1/(6a)$, we obtain 
$$ \sum_{m=1}^{n}{ \sum_{(x_i,x_j)\in B_m}  p(x_i,x_j) \log{\frac{1}{q(\psi_i,\psi_j)}}} \geq  \frac{1}{6a}  \sum_{m=1}^{n}{  \sum_{(x_i,x_j)\in B_m}  \log{\frac{1}{q(\psi_i,\psi_j)}}}.$$
From the monotonicity of $\exp(-x^2)$ we obtain that
$$
\frac{1}{q(\psi_i,\psi_j)} = \frac{  \sum_{k \neq i}{\exp(-\|\psi_i - \psi_k\|^2)}} {\exp(-\|\psi_i - \psi_j\|^2)} \geq    |\psi \cap B(\psi_i, \|\psi_j - \psi_i\|)|
$$
and combines this with the elementary fact 
\st{
$$ \frac{1}{6a} = \frac{1}{a(a-1)} \frac{a-1}{6} \geq  \frac{1}{a(a-1)} \frac{a}{12}  \qquad \mbox{for all}~a \geq 2$$
to write
\begin{align*}
  \frac{1}{6a}  \sum_{m=1}^{n}{  \sum_{(x_i,x_j)\in B_m}  \log{\frac{1}{q(\psi_i,\psi_j)}}} &\geq   \frac{ na }{12} \left[ \frac{1}{n}  \frac{1}{a(a-1)}\sum_{m=1}^{n}{  \sum_{(x_i,x_j)\in B_m}  \log{  |\psi \cap B(\psi_i, \|\psi_j - \psi_i\|)|}} \right] \\
&= \frac{n a}{12} Q(\psi).
\end{align*}
This shows that for any embedding
$$ \frac{na}{12} Q(\psi) \leq L(\psi).$$
By plugging in our computation from above we obtain that for any embedding $\psi^*$ that minimizes the SNE loss 
$$ \frac{na}{12}  Q(\psi^*) \leq L(\psi^*) \leq  6e n a \log{2a} $$
and the result follows.
}
\end{proof}

\subsection{Improved constants.} We observe that for large parameters, we can improve several of the estimates: for every $\varepsilon > 0$, and sufficiently large $n \gg a \gg 1$ (depending only on $\varepsilon$), we can improve the estimate to
$$ L(\psi^*) \leq  \left(2e+ \varepsilon\right) na \log{2a},$$
and furthermore 
$$ p(x_i, x_j) \geq \left( \frac{1}{e} - \varepsilon\right)\frac{1}{a} \qquad \mbox{for all pairs}~x_i,x_j~\mbox{in the same cluster}$$ 
as well as
\st{
$$ \frac{1}{ea} =  \frac{1}{a(a-1)} \frac{a-1}{e} \sim (1-\varepsilon) \frac{1}{a(a-1)} \frac{a}{e}$$
which leads to an overall estimate of 
$$ Q(\psi^*) \leq (2e^2+\varepsilon)\log{2a} \leq 15 \log{2a} .$$
}

\subsection{Constants and the output dimension $d$.} We have avoided emphasizing this point in the statement of the main result for the sake of clarity and because these techniques
are perhaps most often used for $d \in \left\{2, 3\right\}$. However, as the output dimension $d$ increases, we can relax one of the assumptions in the argument. Instead of sending the
cluster $B_i$ to $(i,0, \dots, 0) \in \mathbb{R}^d$, we can map the clusters to a set of lattice points such that the distance between any pair of lattice points is $\sim c_{d} n^{1/d}$ for some
constant $c_d$. Then, however, in the computation of the loss for that particular embedding, we only need to ensure that
$$ e^{-c^2} (c_d^2 n^{2/d} + \log{(2a)})n \lesssim  \log{a},$$
which is achieved for a degree of separation 
$$ c = \min_{x \in B_i, y \in B_j}{\|x -y\|} \geq \sqrt{ \left( 1 + \frac{2}{d} \right) \log{n}}.$$
The final result is then $Q(\psi^*) \lesssim_{d} \log{a}$ for some implicit constant depending on $d$ and the milder separation condition. It seems that some degree
of separation (growing with the number of clusters) is required if we want the result to be independent of the input dimension $D$.

\subsection{The result for general kernels $q$}
\begin{proof}[Sketch of the proof.]  The proof is a relatively straightforward adaption of the previous argument. We start by noting that for the particular embedding sending $B_i$ to the integer $i \in \mathbb{R}$
$$ 
q_2(\psi_i,\psi_j) = \frac{f\left(\|\psi_i - \psi_j\| \right)}{ \sum_{k \neq i} f \left(\|\psi_i - \psi_k\| \right)} \geq \frac{f \left(\|\psi_i - \psi_j\| \right)}{ \sum_{k=-\infty}^{\infty}{a f(|k|) }} \geq  \frac{f\left(\|\psi_i - \psi_j\| \right)}{ 2 \left(\sum_{k=0}^{\infty}{f(k)} \right)a}.
$$ 
By assumption, the sum $\left(\sum_{k=0}^{\infty}{f(k)} \right) < \infty$ and thus
$$
q_2(\psi_i,\psi_j) \gtrsim \begin{cases} f(0)/a \qquad &\mbox{if}~\psi_i~ \mbox{and}~ \psi_j~\mbox{are both in the same cluster} \\  f(n)/a \qquad &\mbox{otherwise.} \end{cases}
$$
Repeating the computation above and using $f(x) \geq \alpha e^{- \beta x^2}$ to ensure that
$$ \log{\left( \frac{1}{q_2(\psi, \psi_j)}\right)} \lesssim n^2 + \log{2a}$$
shows that the specially chosen embedding still satisfies
 $$ L(\psi^*) \lesssim_{f}  na \log{(a)},$$
where the implicit constant depends only on $f$.
As for the lower bound, there is no change in the argument at all since
$$
\frac{1}{q_2(\psi_i,\psi_j)} = \frac{  \sum_{k \neq i}{f(\|\psi_i - \psi_k\|)}} {f(\|\psi_i - \psi_j\|)} \geq    |\psi \cap B(\psi_i ,\| \psi_j - \psi_i\|)|
$$
is valid for any positive and monotonically decreasing function.
\end{proof}


\textbf{Acknowledgements.}
We thank Raphy Coifman, Boaz Nadler and Laurens van der Maaten for helpful discussions.

\bibliography{references}
\bibliographystyle{apalike}

\end{document}